**ARTICLE**

# Focusing on potential named entities during active label acquisition

Ali Osman Berk Şapcı, Hasan Kemik, Reyyan Yeniterzi  and Oznur Tastan

Faculty of Engineering and Natural Sciences, Sabancı University, Istanbul, Turkey
**Corresponding author:** Reyyan Yeniterzi, Oznur Tastan;
E-mails: reyyan.yeniterzi@sabanciuniv.edu, otastan@sabanciuniv.edu

## Abstract

Named entity recognition (NER) aims to identify mentions of named entities in an unstructured text and classify them into predefined named entity classes. While deep learning-based pre-trained language models help to achieve good predictive performances in NER, many domain-specific NER applications still call for a substantial amount of labeled data. Active learning (AL), a general framework for the label acquisition problem, has been used for NER tasks to minimize the annotation cost without sacrificing model performance. However, the heavily imbalanced class distribution of tokens introduces challenges in designing effective AL querying methods for NER. We propose several AL sentence query evaluation functions that pay more attention to potential positive tokens and evaluate these proposed functions with both sentence-based and token-based cost evaluation strategies. We also propose a better data-driven normalization approach to penalize sentences that are too long or too short. Our experiments on three datasets from different domains reveal that the proposed approach reduces the number of annotated tokens while achieving better or comparable prediction performance with conventional methods.

## 1. Introduction

Named entity recognition (NER) aims to identify mentions of named entities in an unstructured text and classify them into predefined named entity classes (e.g., person names, organizations, and locations). NER is a fundamental natural language processing (NLP) task used in other NLP tasks such as entity linking, event extraction, and question answering. Although deep learning-based pre-trained language models (Devlin *et al.* 2019; Yang *et al.* 2019; Liu *et al.* 2019; Raffel *et al.* 2020) have advanced the state-of-the-art performance in NER (Peters, Ruder, and Smith 2019; Torfi *et al.* 2021), a sufficient amount of labeled data is still necessary for achieving satisfactory prediction performance in most domains (Tikhomirov *et al.* 2020; Wang, Mayhew, and Roth 2020). Since acquiring labeled data is costly, efficient label acquisition for NER remains a challenge.

A general framework for tackling the labeled data acquisition problem is active learning, in which the learner strategically chooses the most valuable instances as opposed to selecting a random sample for labeling (Thompson, Califf, and Mooney 1999). Three main active learning settings have been considered in the literature: membership query synthesis, stream-based selective sampling, and pool-based sampling. In the pool-based active learning setup, the active learner selects the most useful examples from an unlabeled pool of samples and queries them to an annotator for labeling; upon receiving the labels for the queried examples, the model is retrained with the augmented labeled set. These query selection, annotation, and retraining steps are iterated multiple times until either the desired performance is achieved or the labeling budget is exhausted (Settles 2011). The goal is to reduce the annotation cost by creating a smaller labeled set while maintaining a good predictive performance. For many NLP tasks, including NER, a large collection of unlabeled data can be gathered easily. Hence, a pool-based scenario is much more common than membership query synthesis or stream-based selective sampling (Settles 2009).

Active learning (AL) has been successful in various sequence annotation tasks such as part-of-speech tagging (Ringger *et al.* 2007), dependency parsing (Li *et al.* 2016b), and semantic parsing (Thompson *et al.* 1999). AL has been used to tackle the label acquisition problem in the NER task as well. Shen *et al.* (2017) demonstrated that AL combined with deep learning achieves nearly the same performance on standard datasets with just 25% of the original training data. Chen *et al.* (2015) developed and evaluated both existing and new AL methods for a clinical NER task. Their results showed that AL methods, particularly uncertainty sampling approaches, provide significant savings in the annotation cost.

In active learning, the most critical step is selecting useful query examples for manual annotation. This step becomes more challenging for sequence labeling tasks, especially for named entity recognition, for two reasons. The first challenge of applying active learning to NER arises due to imbalanced data distribution. In NER annotation, a token is either labeled with its corresponding named entity class if it is part of a named entity or with the "`other`" class if it is not part of a named entity. The `other` class is generally referred to as *negative annotation* or *negative token*, and all other labels—named entity labels—are referred as *positive annotations* or *positive tokens* (Marcheggiani and Artières 2014). In NER datasets, negative tokens are usually over-abundant compared to positive tokens.

The second challenge in applying active learning to NER is caused by the varying length of sentences. In NER, tokens are annotated one by one, but the context, hence the corresponding sentence, is still required for accurate token annotation. Therefore, at each active learning iteration, sentences are queried instead of tokens. Active learners that select the informative sentences for querying by directly aggregating over all the tokens are biased towards longer sentences. In order to prevent this bias toward sentences with more terms, the aggregated sentence scores are normalized by the number of tokens in the sentence (Engelson and Dagan 1996; Haertel *et al.* 2008; Settles, Craven, and Friedland 2008).

This commonly used approach leads to an unintended consequence; this time, the active learner queries sentences that are too short in the early and middle rounds. The end result of choosing short sentences is that there are fewer annotated tokens to train the NER model (Tomanek 2010). In this paper, we moderate these two extreme cases and propose a normalization technique that makes use of the token count distribution in the dataset.

The varying length of sentences also affects the cost evaluation of the active learning framework. Some studies (Kim *et al.* 2006; Settles and Craven 2008; Yao *et al.* 2009; Liu *et al.* 2022) treat all sentences equally and compare active learning methods directly with respect to the number of sentences queried. However, this is not realistic since the cost is not fixed across sentences, as sentences differ in the number of tokens and the number of named entities they contain (Haertel *et al.* 2008; Arora, Nyberg, and Rosé 2009). Therefore, the number of annotated tokens should be incorporated into the active learning cost. In this regard, many studies in the literature (Shen *et al.* 2004; Settles and Craven 2008; Reichart *et al.* 2008; Shen *et al.* 2017) measure the cost of the annotation by the number of tokens annotated even though they query the sentences. Using only the token count is also an imperfect strategy, as the cost of annotating the same number of tokens distributed over multiple sentences is not equivalent to annotating these tokens within a single sentence (Settles, Craven, and Friedland 2008; Tomanek and Hahn 2010). This is mainly because there is a cost factor associated with each new sentence that is independent of its content and length. Even though we do not propose a new cost calculation method that encompasses all these different aspects, we consider these two cost evaluation setups to analyze the existing and proposed approaches in detail.

**Table 1.** Details and statistics of the datasets.

| Dataset | Total # Sentences[a] | Total # Tokens[b] | Average # Token[c] | Average # Pos. Token[d] | Percentage Pos. Token[e] |
|---------|---------------------|-------------------|--------------------|------------------------|--------------------------|
| CoNLL-03 | 17,291 | 254,983 | 14.75 | 2.45 | 17% |
| BC5CDR | 9357 | 242,920 | 25.96 | 3.03 | 12% |
| NCBI-Dz. | 6372 | 160,583 | 25.20 | 2.08 | 8% |

[a]Total number of sentences.
[b]Total number of tokens.
[c]Average number of tokens per sentence.
[d]Average number of positive tokens per sentence.
[e]Percentage of positive tokens to all tokens.

In this study, we propose an extension to the existing uncertainty sampling methods to handle the challenges associated with the overabundance of negative tokens. In our proposed approach, the query evaluation metrics are designed to focus less on the tokens that are predicted to have negative annotations. We identify potentially negative tokens through clustering of pre-trained BERT representations using a semi-supervised dimensionality reduction step. Using BERT embeddings directly in the active learning querying step for NER is a novel contribution. Finally, this paper proposes a normalization strategy for aggregating token scores to attain a better sentence query metric that favors sentences that are not too short or too long. For a fair comparison, we evaluate different active learning query methods, both under the assumption of fixed annotation cost per sentence and fixed annotation cost per token. Our experiments on three datasets from different domains illustrate that our proposed approach reduces the number of annotated tokens while maintaining a slightly better or equal prediction performance to the compared methods. We also present an extensive investigation of the effects of different pre-trained language embeddings on the performance of our NER model.

The rest of the paper is organized as follows: Section 2 presents the NER data collections used in the experiments together with additional details to motivate the reader for the proposed method, Section 3 summarizes the general active learning setup and the commonly used active learning strategies for the NER task, and Section 4 describes the proposed approach. We describe our experimental setting in Section 5 and detail our results in Section 6. Section 7 concludes with a summary of our findings.

## 2. Datasets

Three commonly used English NER datasets from two different domains are selected.

- CoNLL-03 (Tjong Kim Sang and De Meulder 2003): This dataset contains four different named entities, namely `Person`, `Organization`, `Location,` and `Miscellaneous` with the following percentages 33%, 28%, 24%, and 14%, respectively.
- BC5CDR (Li *et al.* 2016a): BioCreative V CDR dataset contains equal distribution of `Disease` and `Chemical` tags.
- NCBI-Disease (Doğan, Leaman, and Lu 2014): This dataset only contains the `Disease` tag.

Some important statistics on these datasets are provided in Table 1. As it can be seen from Table 1, the ratio of the positive tokens (annotations) is generally low. CoNLL-03, which has four different named entities, contains the highest percentage with 17%, and this ratio is 8% in NCBI-Disease when there is only one type of named entity. This imbalance causes the problems that are detailed in Section 1. We propose to use the predicted annotation type of tokens in the candidate selection setting to address this problem.

## 3. Active learning for NER

Let $\mathbf{x} = \langle x_1, x_2, \cdots, x_n \rangle$ be a sequence of tokens in a sentence. Let $\mathbf{y} = \langle y_1, y_2, \cdots, y_n \rangle$ be the sequence of named entity states, each of which is associated with a named entity type, $y \in \{y_1, y_2, \cdots y_m, o\}$, where $m$ is the number of different named entity classes and $o$ for the `other` class. In the NER task, our aim is to uncover the hidden sequence of $\mathbf{y}$ given the observed sequence of $\mathbf{x}$.

In the pool-based AL, the pool consists of unlabeled sentences, and at each iteration, the AL strategy aims to identify the most useful examples to be annotated. Since named entities often span multiple tokens and being a named entity or not depends on the context, the fundamental annotation unit in NER is not a token but a sentence.

Thus, in the AL framework, the unlabeled pool, $U$, contains the unlabeled sentences, and the labeled set, $L$, contains the annotated sentences. At each iteration, a set of sentences is selected as the query set from $U$, and upon annotation, added to $L$. Overall, the active learning loop proceeds as follows:

- **Initial Model Creation:** Initially, some sentences from the set $U$ are selected randomly. These sentences, which are referred to as $q_0$, are manually annotated and included to set $L$. Then, the first NER model is trained with this initial annotated set.
- **Iteration:** The following steps are repeated until the chosen stop criterion is met.

  - **Querying:** In the query step at iteration $j$, examples in the $U$ are ranked based on the querying metric and, the top-ranked $|q_j|$ sentences are selected as the query batch.
  - **Annotation:** The selected set of sentences, $q_j$, are manually annotated and added to the set $L$.
  - **Training:** The NER model is retrained on the augmented labeled set $L$.

The querying step is the most critical step in AL since the most useful and informative examples for learning are chosen at this step of annotation. One of the most common general frameworks for measuring informativeness is uncertainty sampling (Lewis and Catlett 1994), where the active learner selects the unlabeled example(s) for which the model at the time is most uncertain about its prediction. In our case, the active learner ranks the sentences based on the uncertainty measure, then the top-ranked sentences are queried as a batch for annotation. Since sentences are annotated at NER, an uncertainty score $\Phi$ is calculated at the sentence level. $\Phi(\mathbf{x}^*) > \Phi(\mathbf{x}')$ indicates that the model at the time is more uncertain about $\mathbf{x}^*$. Thus, it will prefer $\mathbf{x}^*$ over $\mathbf{x}'$ when querying.

Various strategies exist to estimate a token's uncertainty, and also different ways are available to calculate the final sentence uncertainty $\Phi$ (Settles and Craven 2008; Haertel *et al.* 2008; Tomanek 2010). There are two main ways to assess the informativeness/usefulness of a sentence in AL. One can calculate a score over the entire sentence based on the conditional probability of the Viterbi sequence given the observation sequence.

Alternatively, the score of a sentence can be calculated by aggregating token-level scores, which are calculated at each position $i$ of the observation sequence by the marginal probability of the respective labels. A detailed comparison of several utility functions indicates that one is not necessarily superior to the other. Yet, token aggregated methods perform slightly better and have lower computational complexity, since only the best Viterbi sequence needs to be determined (Tomanek 2010).

After querying the most informative sentences, their labels are acquired. In the next step, the NER model, in our case a CRF model, is retrained with the augmented labeled sentence set. Since the step of acquiring labels usually comes with a cost, a cost evaluation metric is used to determine the cumulative cost of active learning iterations. Depending on the stopping criterion, the measured total cost may lead to the termination of the active learning loop. Likewise, the performance of the retrained NER model can be evaluated on a test set to assess the generalization performance of the model. Alternatively, evaluated performance can also be used to determine if the stopping criterion is met, together with the total annotation cost.

### 3.1 Assessing uncertainty of tokens

The uncertainty score of a sentence is calculated by aggregating the tokens' uncertainties (Shen *et al.* 2004; Settles and Craven 2008; Chen *et al.* 2015; Liu *et al.* 2022). Therefore, estimating token uncertainties is critical in assessing the value of a sentence. In this subsection, we will review four different approaches for measuring the uncertainty of a token.

- **Entropy-based:** A common way to quantify the class labels' uncertainty is to calculate the entropy (Shannon 1948) of class labels. For the NER task, the entropy of the class label for the token, $x_i$, is denoted as $H(x_i)$ and is computed over all possible tags as follows:

$$H(x_i) = -\sum_{j=1}^{M} P(y_i = j | \mathbf{x}, \theta) \log P(y_i = j | \mathbf{x}, \theta) \tag{1}$$

  where $\theta$ represents the model parameters, $y_i$ is the predicted label for the token $x_i$, and $M$ is the number of possible class labels for a token. $P(y_i = j | \mathbf{x}, \theta)$ stands for the estimated probability of tag $j$ being the label of the token $x_i$.

- **Token probability:** The uncertainty associated with the most likely tag assignment of the token is another measure of uncertainty. Let $T(x_i)$ be the token probability, which is computed based on the most probable entity label's estimated probability for $x_i$ in $\mathbf{x}$ as follows:

$$T(x_i) = 1 - \max_{1 \le j \le M} P(y_i = j | \mathbf{x}, \theta). \tag{2}$$

- **Assignment probability:** Alternatively, instead of considering each token in isolation, we can consider the most likely label assignment of the sentence, which we will refer to as $\hat{\mathbf{y}}$, Viterbi sequence, and then use the assigned probability of the token. The uncertainty for a token $x_i$ in that sequence assignment is computed based on the probability of the label assignment $\hat{y}_i$ of $x_i$ under the most likely tag sequence $\hat{\mathbf{y}}$ (Liu *et al.* 2022). We call this measure as the assignment probability, denoted by $A(x_i)$, is given by the following equation:

$$A(x_i) = 1 - P(y_i = \hat{y}_i | \mathbf{x}, \theta). \tag{3}$$

- **Margin-based:** Another metric of uncertainty is the margin between the most and the second most probable entity labels of the token $x_i$ (Scheffer, Decomain, and Wrobel 2001). $M(x_i)$ is defined as follows:

$$M(x_i) = 1 - \left( \max_{1 \le j \le M} P(y_i = j | \mathbf{x}, \theta) - \max_{\substack{1 \le k \le M \\ k \ne j}} P(y_i = k | \mathbf{x}, \theta) \right) \tag{4}$$

  where $j$ and $k$ are the most likely and the second most likely label assignments for the token $y_i$.

These four metrics, $H(x_i)$ token entropy, $T(x_i)$ token probability, $A(x_i)$ assignment probability, and $M(x_i)$ token margin constitute the four alternative token uncertainty measures we used throughout this paper. We will refer to the token uncertainty measure as $\tau$ in the upcoming sections.

### 3.2 Assessing uncertainty of sentences

Tokens' uncertainty scores are used to evaluate sentence informativeness for querying. Several methods have been proposed in the literature in order to aggregate the token scores into a single sentence score. Three of these, namely *Single Most Uncertain* (`single`), *Normalized Sum* (`normalized`), and *Total Sum* (`total`), will be described before detailing our proposed approach.

- **Single Most Uncertain:** One way of deriving the uncertainty score of a sentence $\mathbf{x}$ is considering the token $x_i \in \mathbf{x}$ with maximum uncertainty value $\tau(x_i)$ as follows:

$$\Phi_s(\mathbf{x}) = \max_{1 \leq i \leq N_{\mathbf{x}}} \tau(x_i) \tag{5}$$

  where $N_{\mathbf{x}}$ is the total number of tokens in the sentence $\mathbf{x}$.

- **Normalized Sum:** The previous method focuses on the most uncertain one, but ignores the other tokens' usefulness in the sentences. An alternative method is to sum the uncertainty scores of tokens and normalize this sum with the number of tokens in the sentence, $N_{\mathbf{x}}$ (Hwa 2004; Baldridge and Osborne 2004).

$$\Phi_n(\mathbf{x}) = \frac{1}{N_{\mathbf{x}}} \sum_{i=1}^{N_{\mathbf{x}}} \tau(x_i) \tag{6}$$

- **Total Sum:** Settles and Craven (2008) prefer to use aggregated values directly without normalizing with the token count, as longer sentences tend to contain more information.

$$\Phi_t(\mathbf{x}) = \sum_{i=1}^{N_{\mathbf{x}}} \tau(x_i) \tag{7}$$

## 4. Proposed active learning querying approach

In this paper, we propose a new sentence query evaluation function that considers the imbalance between the positive and negative tokens and the length of the query sentences. Our proposed evaluation function requires identifying likely positive tokens at each query step, which we solve through a semi-supervised clustering approach on token embeddings. In the following sections, we will detail this approach and the proposed normalization strategy based on sentence lengths.

### 4.1 Positive token biased querying

As shown in Table 1, negative tokens are far more abundant in sentences than positive tokens (Tjong Kim Sang and De Meulder 2003). This imbalance challenges the AL query evaluation step; whenever the sentence's overall uncertainty is calculated, the negative tokens' uncertainty scores dominate the overall score, and shadow the positive tokens. This causes the trained prediction model to miss the opportunity of using sentences with informative, positive tags that are very useful to identify whether a term is a named entity or not and if it is, the correct type of it. We propose to solve this problem by having a query function that focuses more on the positive tokens. However, the information of which tokens are positive is unknown at the query time, and thus, it needs to be guessed. For this, we first identify these likely positive tokens using a semi-supervised clustering approach. To achieve this, we exploit pre-trained language model embeddings directly in the querying step.

### 4.1.1 Identifying possibly positive tokens

Contextual word embeddings provide significant improvements in many NLP tasks, mainly because these models are pre-trained over large data collections that enable them to capture the different characteristics of terms depending on the context (Liu, Kusner, and Blunsom 2020). Domain-specific pre-trained models, such as BioBERT, even provide more useful domain-enriched representations. In this paper, we use these BERT representations to identify the possibly positive tokens. Since terms that have similar semantic and syntactic roles are expected to have similar embeddings, we resort to clustering in order to identify the possible groups. Based on our observation of the datasets and distribution of tags among these clusters, we assume that the largest cluster will be the negative token cluster. The rationale behind this assumption is that the imbalanced distribution of the data shall reveal itself in the sizes of these clusters; therefore, the largest cluster should be the negative one. Based on this assumption, we consider all the tokens in other clusters as the positive token set. The final clusters may still contain some outliers. Especially, the tokens that are in the largest cluster but away from the corresponding cluster centroid might also be informative and, therefore, desirable to query. Thus, we extend the positive token set with tokens that are scattered away from cluster centroids and identified as outliers.

Several adaptations and extensions are needed to apply this idea. Firstly, contextual embeddings, like BERT or BioBERT, are high dimensional. When the embedding space is high-dimensional, clusters tend to split in a sparse space. When that is the case, one cannot confidently assume that the largest cluster will consist of primarily negative tokens because tokens are all grouped into small clusters with similar sizes. To prevent this, we propose reducing the original embeddings to a lower-dimensional space and performing clustering in this lower-dimensional space.

Another problem with the existing approach can arise due to the fact that both the dimensionality reduction and the clustering are unsupervised algorithms. Using unsupervised approaches may result in unexpected clusters due to the unrelated characteristics of the tokens. To avoid this, we guide the clusters using task-specific labels. Instead of using an unsupervised dimensionality reduction, we use a semi-supervised approach that makes use of the subset of available annotations at the query time. Along with the active learning iterations, the labeled set grows as well, which results in better dimensionality reduction performance over iterations.

For the clustering step, we use the Hierarchical Density-Based Spatial Clustering of Applications with Noise (HDBSCAN) algorithm (Campello, Moulavi, and Sander 2013; Campello *et al*. 2015). It has been shown that Uniform Manifold Approximation and Projection (UMAP) (McInnes, Healy, and Melville 2020) can be used as an effective pre-processing step to boost the performance of density-based clustering algorithms. We use the semi-supervised extension of UMAP (Sainburg, McInnes, and Gentner 2021). For detecting tokens that are close to the cluster boundaries, we use outliers provided by the Global-Local Outlier Score from Hierarchies (GLOSH) algorithm (Campello *et al*. 2015). Tokens with the top 1% outlier scores are used.

To summarize, we execute the following steps to obtain a set of likely positive tokens:

1. We obtain the contextualized word embeddings of the tokens that are in the union of the labeled and unlabeled token set, $U \cup L$, using a pre-trained model. These pre-trained models are domain-specific whenever available.

2. The embeddings of all tokens in $U \cup L$ are mapped into a two-dimensional space using semi-supervised UMAP. Tokens in $L$ are used for supervision.

3. HDBSCAN is used to cluster and identify possibly positive tokens. The largest cluster is assumed to be the negative token cluster, and the remaining clusters are merged to form the possibly positive token set $P$.

4. We include the topmost %1 outlier tokens based on the outlier scores provided by the GLOSH algorithm. Let $T$ be the most %1 outlier tokens, we define the extended positive token set such that $P' = P \cup T$.

Step 1 is performed only once, but steps 2–4 are executed at each active learning iteration. A token $x$ is estimated to have a positive annotation if and only if $x \in P'$, which is constructed as described above.

### 4.1.2 Total positive sum score

Here we propose to use only the possibly positive ones instead of all tokens. Therefore, the $P'$ set which is predicted at the active learning iteration at the time, is used. This approach, which we call as the *Total Positive Sum* (`total-pos`), is shown below:

$$\Phi_{tp}(\mathbf{x}) = \sum_{\substack{1 \le i \le N_{\mathbf{x}} \\ x_i \in P'}} \tau(x_i). \tag{8}$$

Here, $P'$ are the set of tokens that are estimated to be positively annotated. This is very similar to Equation (7). The only difference is that this one performs the calculations over a subset of tokens (possibly positive ones), not all

### 4.2 Density normalization

The second extension that we propose promotes the selection of sentences with typical length in the distribution. When calculating a query metric for a sentence, if no normalization is applied, the aggregated token score favors longer sentences (Culotta and McCallum 2005; Tomanek 2010). The most common way to prevent this is to normalize the result with the number of tokens in the sentence (Hwa 2004; Baldridge and Osborne 2004; Shen *et al.* 2017). This on the other hand leads the AL method to choose too short sentences (shown in Section 6). This is also not desirable as choosing shorter sentences requires more iterations to exceed a predetermined performance level and therefore more manual labeling. Therefore, while querying extremely long sentences should be avoided, querying too short sentences should be avoided as well.

The second modification we propose deals with this issue. We would like to query sentences that are not too short or too long compared to other sentences in the dataset. For this purpose, we directly use the dataset's distribution of the token count and sample a sentence in proportion to its token count frequency.

### 4.2.1 Density normalized positive sum score

Our second proposed strategy is the uncertainty over the tokens estimated to have *positive annotations* and favors sentences with an average token count. This strategy, called *Density Normalized Positive Sum Score* (`dnorm-pos`), is computed as follows:

$$\Phi_{dp}(\mathbf{x}) = \sqrt{p_L(N_{\mathbf{x}})} \sum_{\substack{1 \le i \le N_{\mathbf{x}} \\ x_i \in P'}} \tau(x_i), \tag{9}$$

where $p_L$ denotes the probability density function of the token counts of sentences in the dataset, which is estimated by Gaussian kernel density estimation (Silverman 1982). The evaluation metric $\Phi_{tp}(\mathbf{x})$ is modified by multiplying with $\sqrt{p_L(N_{\mathbf{x}})}$. This way, sentences are ranked based on the values of $\sqrt{p_L(N_{\mathbf{x}})}\Phi_{tp}(\mathbf{x})$, and hence, sentences with more probable token counts are preferred.

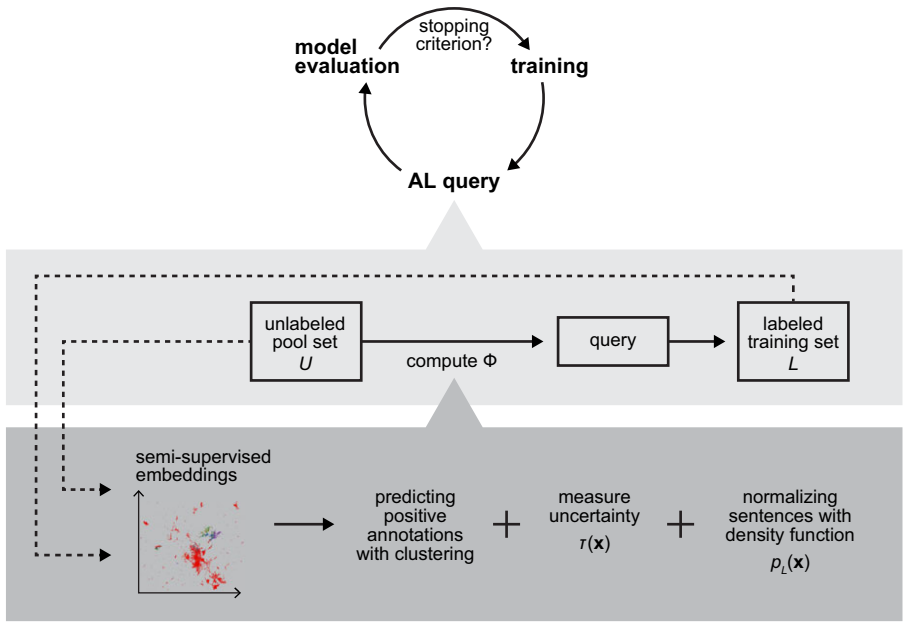

**Figure 1.** Summary of the proposed active learning framework for the NER task. The main active learning loop consists of three steps (model evaluation, training, and active learning query) and continues until the stopping criterion is satisfied. For each query, uncertainties of unlabeled sentences are estimated with function Φ. Then, the most uncertain sentences are sent to an annotator to expand the labeled training set. To compute Φ, we propose to focus on tokens that are predicted to have positive annotations. Our approach for predicting positive tokens consists of computing semi-supervised embeddings of tokens and density-based clustering.

### *4.3 Proposed active learning query scheme*

Figure 1 summarizes the proposed active learning query scheme as a whole. As explained in Section 3, an active learner queries the annotator based on the sentence score Φ at each iteration. Both *Density Normalized Positive Sum Score* and *Total Positive Sum Score* approaches require constructing the positive token set $P'$. As the labeled set $L$ is updated at each iteration, the semi-supervised embeddings are updated with the labeled sentence set $P'$ is computed at each iteration as described in Section 4.1.1. Subsequently, the *Total Positive Sum Score* is calculated using the desired uncertainty measure $\tau$ as shown in Section 4.1.2. Finally, density normalization is applied as explained in Section 4.2.

## 5. Experimental setting

### *5.1 Baseline methods*

The proposed uncertainty-based querying methods are compared to the following commonly used baselines and strong baselines proposed herein:

- **Random Selection (RS):** samples sentences uniformly at random from the unlabeled pool.
- **Longest Sentence Selection (LSS):** is a simple baseline that selects the sentence with the highest number of tokens.
- **Positive Annotation Selection (PAS):** While RS and LSS are standard baselines for AL methods, we propose PAS as a strong baseline for the proposed query function. PAS is similar to the proposed function, except it ignores the uncertainty of the tokens. It queries based on the number of likely positive tokens and the number of tokens in the sentences.

**Table 2.** Uncertainty-based querying methods and their abbreviations used throughout the text.

|  | Single most uncertain (`single`) | Normalized sum (`normalized`) | Total sum (`total`) |
|---|---|---|---|
| Assignment | sAP | nAP | tAP |
| Probability | Single Assignment Probability | Normalized Assignment Probability | Total Assignment Probability |
| Token | sTE | tTE | nTE |
| Entropy | Single Token Entropy | Normalized Token Entropy | Total Token Entropy |
| Token | sTP | nTP | tTP |
| Probability | Single Token Probability | Normalized Token Probability | Total Token Probability |
| Token | sTM | nTM | tTM |
| Margin | Single Token Margin | Normalized Token Margin | Total Token Margin |

See Sections 3.2 and 5.2 for definitions.

We define PAS as follows:

$$\text{PAS}(\mathbf{x}) = \sqrt{p_L(N_\mathbf{x})} \sum_{i=1}^{N_\mathbf{x}} \mathbb{1}_{p'}(x_i) \tag{10}$$

where $x_i$ is the $i$th token of sentence $\mathbf{x}$ and $N_\mathbf{x}$ is the number of tokens in the $\mathbf{x}$.

### 5.2 Uncertainty-based querying methods

In addition to these simple baselines, 12 querying strategies which are constructed by combining the 4 token uncertainty measures with 3 sentence uncertainty measures are used in the experiments. AP, TE, TP, and TM stand respectively for the different values of $\tau$: $A$, $H$, $T$, and $M$. These approaches are listed in Table 2.

The calculation of two simple baselines (LSS and PAS) and other sentence uncertainty scores are further demonstrated in the example sentence "Angioedema due to ACE inhibitors: common and inadequately diagnosed" (Figure 2). The last two lines in Figure 2 shows the proposed aggregated uncertainty measures, $\Phi_{tp}$ for total positive sum score and $\Phi_{dp}$ for density normalized positive sum score. These proposed sentence scores are computed for different token uncertainty measures that are listed in Table 3.

### 5.3 NER model

We use Conditional Random Fields (CRF) (Lafferty, McCallum, and Pereira 2001) as the NER prediction model to evaluate the different active learning strategies. Combining representations from large pre-trained models with structured predictions of CRF (Lafferty *et al.* 2001) has been shown to achieve good performance in the NER task when labeled data are not abundant (Souza, Nogueira, and Lotufo 2020). Souza *et al.* (2020) achieve state-of-the-art performance on a popular Portuguese NER corpus by pre-training a Portuguese BERT model on a large unlabeled corpus, and a CRF model is trained with frozen weights of the BERT model for the Portuguese NER task. In this paper, we follow their approach and use BERT representations as features in the CRF model. But, instead of pre-training a language model for different domains, we employ appropriate publicly shared pre-trained BERT models, as described in Section 5.3.1.

**Table 3.** Proposed aggregated uncertainty measures for each token uncertainty measure.

|  | Total positive sum (`total-pos`) | Density normalized positive sum (`dnorm-pos`) |
|---|---|---|
| Assignment | tpAP | dpAP |
| Probability | Total Positive Assignment Probability | Density Normalized Positive Assignment Probability |
| Token | tpTE | dpTE |
| Entropy | Total Positive Token Entropy | Density Normalized Positive Token Entropy |
| Token | tpTP | dpTP |
| Probability | Total Positive Token Probability | Density Normalized Positive Token Probability |
| Token | tpTM | dpTM |
| Margin | Total Positive Token Margin | Density Normalized Positive Token Margin |

See Sections 4.1.2, 4.2.1 and 5.2 for definitions.

(1)  $\mathbf{x} := [[\text{Angioedema}]_B]_{\text{Disease}}$ due to $[[\text{ACE}]_B \ [\text{inhibitors:}]_I]_{\text{Chemical}}$ common and inadequately diagnosed.

(2)  [Angioedema] [due] [ to ] [ACE] [inhibitors:] [common] [and] [inadequately] [diagnosed]

| [ $x_1$ ] | [ $x_2$ ] | [ $x_3$ ] | [ $x_4$ ] | [ $x_5$ ] | [ $x_6$ ] | [ $x_7$ ] | [ $x_8$ ] | [ $x_9$ ] |
|---|---|---|---|---|---|---|---|---|

$$\text{LSS}(\mathbf{x}) = \sum_{i=1}^{9} 1 = 9 \qquad \text{PAS}(\mathbf{x}) = \sqrt{p_L(9)} \sum_{i=1}^{9} \mathbb{1}_{P'}(x_i) = 3\sqrt{p_L(9)}$$

$$\Phi_s(\mathbf{x}) = \max_{1 \le i \le 9} \tau(x_i) \qquad \Phi_n(\mathbf{x}) = \frac{1}{9} \sum_{i=1}^{9} \tau(x_i) \qquad \Phi_t(\mathbf{x}) = \sum_{i=1}^{9} \tau(x_i)$$

$$\Phi_{tp}(\mathbf{x}) = \sum_{\substack{1 \le i \le 9 \\ x_i \in P}} \tau(x_i) = \tau(x_1) + \tau(x_4) + \tau(x_5)$$

$$\Phi_{dp}(\mathbf{x}) = \sqrt{p_L(9)} \sum_{\substack{1 \le i \le 9 \\ x_i \in P}} \tau(x_i) = \sqrt{p_L(9)} \ (\tau(x_1) + \tau(x_4) + \tau(x_5))$$

[*] Assuming positive annotations are estimated correctly.

**Figure 2.** An example sentence and corresponding computations of different query evaluation functions[*].

### 5.3.1 BERT representations

For the CoNLL-03 corpus, word embeddings are extracted from the BERT-base cased model (Devlin *et al.* 2019). Since BC5CDR and NCBI-Disease datasets are more domain-specific, models trained on data of similar domains can be more effective. Therefore, we compare different domain-specific BERT models, namely BioBERT v1.1 cased (Lee *et al.* 2020) and Clinical-BioBERT cased (Alsentzer *et al.* 2019) in a passive learning setting. BioBERT cased v1.1 performs better on both NCBI-Disease and BC5CDR. As expected, BioBERT v1.1 also achieves significantly higher $F_1$-scores ($+\sim 4$) than using BERT in BC5CDR and NCBI-Disease datasets. Therefore, we used BioBERT for both BC5CDR and NCBI-Disease datasets.

There are alternative ways of extracting word embeddings from a BERT model as features (Devlin *et al.* 2019). We compare three alternatives:

- **LL**: Use the last layer directly. The number of hidden units of each layer in BERT equals 768. Therefore, the final representation of each token is also a vector of dimension 768.
- **SL4**: Sum the last 4 layers. Again, each token is represented with a vector of dimension 768.

**Table 4.** Comparison of approaches for extracting BERT embeddings based on passive learning $F_1$-scores.[a] First header row indicates the dimension of embeddings. The second header row stands for the strategies used to obtain embeddings from BERT.

|         | 3072 | 768 | 768 | reduced-300 | | | reduced-256 | | | reduced-200 | | |
|---------|------|-----|-----|------|------|------|------|------|------|------|------|------|
| Dataset | CL4[b] | SL4[c] | LL[d] | CL4 | SL4 | LL | CL4 | SL4 | LL | CL4 | SL4 | LL |
| CoNLL-03 | 92.7 | 92.2 | 92.7 | 93.3 | 71.6 | 92.7 | **93.6** | 65.6 | 92.9 | 93.2 | 60.7 | 92.7 |
| BC5CDR | 85.7 | 85.2 | 85.3 | 85.9 | 85.8 | 85.1 | 85.7 | 85.8 | 85.0 | **86.0** | 85.5 | 85.1 |
| NCBI-Dz. | 82.2 | 80.5 | 82.6 | 84.0 | 82.4 | 82.2 | 84.4 | 84.2 | 82.4 | **84.5** | 83.6 | 83.5 |

[a]For BC5CDR and NCBI-Dz. BioBERT cased v1.1, for CoNLL-03 BERT cased base.
[b]Concatenate last four layer.
[c]Sum last four layer.
[d]Use last layer directly.

- **CL4**: Concatenate the last 4 layers. This one results in a 768·4=3072 dimensional vector for each token.

Since we observe that the performances of the different strategies tend to depend on the dataset, we report active learning experiment results through the best-performing strategy on the corresponding test data (see Table 4).

### 5.3.2 CRF classification

For token-level classifications, we use a Linear-Chain CRF. For an input sentence $\mathbf{x} = \langle x_1, x_2, \cdots, x_n \rangle$ and corresponding sequence of predictions $\mathbf{y} = \langle y_1, y_2, \cdots, y_n \rangle$, the modeled probability distribution $p(\mathbf{x}|\mathbf{y})$ has the following form:

$$p(\mathbf{x}|\mathbf{y}) \propto \prod_{i=1}^{N_{\mathbf{x}}} \exp \left\{ \sum_{k=1}^{K} \lambda_k f_k(y_{i-1}, y_i, x_i) \right\} \tag{11}$$

where $K$ is the total number of features, $f_k$ is the feature function, and $\lambda_k$ is the corresponding feature weight.

CRF features of a single token consist of the following values for the token itself, the previous token, and the next token: the N-dimensional BERT embedding vector, POS-tag, whenever available,[a] the lower-cased token, the last three and the last two letters of the token, and a boolean value for each of the following checks: title, digit, and lower-case. Thus, each feature dictionary consists of $(N+7)\cdot3 = 3N+21$ values.[b]

Training time scales linearly with the average number of features that goes into the token classifier. In our case, the average number of features, $N$, is the size of the BERT embeddings, and it is the dominating term since it can be either 768 or 3072. To reduce the run time, we project BERT embeddings to a lower-dimensional space using principal component analysis (PCA) (Pearson 1901). We choose the number of principal components such that they collectively explain at least ~80%–~85% of the variance.

We optimize the feature weights ($\lambda_k$) using L-BFGS (Liu and Nocedal 1989), a limited-memory variation of the popular and robust BFGS algorithm. Due to its linear memory requirement and polynomial complexity, the L-BFGS algorithm is particularly suitable for our case. The model is trained repeatedly with hundreds of features at each AL iteration. CRFsuite (Okazaki 2007) implementation is used for the experiments. Our Linear-Chain CRF model is configured by setting the maximum number of iterations of L-BFGS to 100, the epsilon parameter that determines the condition of convergence to $10^{-5}$, the period, and the threshold parameters (delta) for the stopping criterion to 10 and $10^{-5}$, respectively. $L_1$ and $L_2$ regularization coefficients ($c_1$ and $c_2$) are both set to 0.1. The line search algorithm used in L-BFGS updates is *More and Thuente's method*, where the maximum number of trials is set to 20. We allow our model to generate negative transition features and negative state features that do not occur in the training data.

---

[a]Only CoNLL-03 contains parts of speech tags. BC5CDR and NCBI-Disease do not involve POS tags.
[b]$(3N+6)\cdot3 = 3N+18$ for the datasets which do not contain POS tags.

We use the standard split of training/validation/test sets of the datasets with the following percentages; 68%–16%–16% for CoNLL-2003, 33%–33%–33% for BC5CDR, and 75%–12.5%–12.5% for NCBI-Disease. Validation sets are not used for training, they are only used to determine hyper-parameters such as the number of iterations, training configuration of the CRF model etc.

### 5.3.3 NER experiments with the entire dataset

As explained above, NER setting consists of several choices like how to use the embeddings or the target dimensions for the dimensionality reduction step. In order to find the right setting for each dataset, different possibilities are explored initially using the entire dataset. In reality, of course, the entire labeled set is not accessible to the active learner. However, since our aim is to investigate the performance of each AL query evaluation function, making the right decisions in other parts of the system will help us to better evaluate the methods. Therefore, we report AL experiments using the embeddings resulting in the highest passive learning $F_1$-score on the test set. In addition to the three types of embedding combinations (LL, SL4, and CL4), the arbitrarily chosen following three values 300, 256, and 200 are tried as possible target dimensions for PCA. The results of the experiments are summarized in Table 4.

According to Table 4, models trained with reduced dimensions improve the predictive performance in many cases. Using the original high-dimensional representation obtained by concatenating the last four layers of BERT (CL4) resulted in a 92.7 $F_1$-score for the CoNLL-2003 dataset, whereas reducing the 3072 dimensional representations to 256 dimensions improved the overall performance by ∼0.9 points. Similarly, reducing the 3072 dimensional original representations of CL4 to 200 dimension increases the $F_1$-score by 2.3 and 0.3 respectively for NCBI-Disease and BC5CDR datasets. For each of the three datasets, the CL4 representation consistently outperforms both the SL4 and LL in reduced forms.

These experiments guide us to choose the optimum NER setting for each dataset. We perform active learning experiments by using the model achieving the highest $F_1$-score on the test split for the corresponding dataset.

### 5.4 Active learning setup

For the active learning task, the classical setup described in Section 3 is used. For the initial model setup, $2^m$ sentences are selected from the set $U$ randomly, where $m$ is an integer that depends on the input dataset. These sentences ($q_0$) are first included in the set $L$, and the first NER model is trained with this initial annotated set. All the compared methods start with the same initial label set.

During the querying step, increasing the batch size is a common practice since it becomes harder to observe a performance improvement at later stages (Guyon *et al.* 2011). We define the batch size at iteration $j$ as $|q_j| = 2^{j+m}$.

Each experiment is repeated nine times with randomly selected initial labeled sets. Since initial NER models may affect the performances of active learning methods, we report the average $F_1$-scores over these nine runs. Implementation of active learning framework and methods together with experiment scripts and configuration files are all publicly available.[c] All results can be reproduced with the shared configuration files.

---

[c] https://github.com/bo1929/anelfop

## 6. Results

This section will summarize the experimental results of different active learning query strategies on three NER datasets. In evaluating the query strategies, we use two different cost measures. As the entire sentence is sent for annotation and all tokens in the sentence are annotated at once, one way of assessing the annotation cost is to directly use the number of sentences queried (Kim *et al.* 2006; Settles and Craven 2008; Yao *et al.* 2009; Liu *et al.* 2022). When a single sentence is queried at each iteration, the number of annotated sentences is equivalent to the number of active learning iterations. Although this sentence-based cost calculation acknowledges the fact that sentences are reviewed at once, it does not take into account that different sentences contain different number of tokens (Haertel *et al.* 2008; Arora *et al.* 2009). An alternative strategy is to measure the cost of labeling with the number of tokens annotated (Shen *et al.* 2004, 2017; Settles and Craven 2008; Reichart *et al.* 2008). In this paper, we use both sentence-based and token-based cost evaluation to get a deeper insight into the merits and weaknesses of different active learning query strategies.

We compare three baseline methods and twenty active learning querying strategies. Among the three standard baseline methods, one of them, the positive annotation selection (PAS) approach, is also proposed herein. Among the twenty strategies deployed, in this study, eight are novel. The remaining twelve are strong baselines against which the eight proposed methods are compared. The $F_1$-scores of all methods compared at the same number of annotated sentences are available in Appendix S1. We summarize these results in Table 5, by taking the mean of the four token uncertainty measures: assignment probability (AP), token entropy (TE), token probability (TP), and token margin (TM), for each sentence level uncertainty score.

### 6.1 Comparison of different baselines

The first three rows in Table 5 list the performances for the three baselines, the middle three rows display active learning query strategies used previously for NER, and the last two rows present the results of the proposed AL strategies. In this table, for each dataset, $F_1$-score values from the last four iterations before convergence are shown, and performances are compared at the same cost, where cost is measured by the number of sentences annotated. The first thing we notice is that in most cases, all query methods outperform baselines. There are few exceptions; for example, `normalized` method can fall behind baselines in the early rounds of AL iterations. This could be because the `normalized` approach chooses shorter sentences that result in a much fewer number of annotated tokens to train with when the same number of sentences are annotated.

Among the three baselines (RS, LSS, and PAS), there is not a clear winner that outperforms all other baselines consistently. One unexpected behavior we observe is that RS consistently performs better than LSS and PAS in early iterations across all datasets. We expect LSS to be better, especially in early rounds, since it selects longer sentences and more tokens will be annotated at early iterations compared to other approaches. However, this expectation does not turn out to be true. The reason could be that in these datasets, long sentences may not contain any positive tokens useful for the classification. To understand the behavior of these baselines more, we examine the results with the second cost measure, the number of annotated tokens. Figure 3 compares the $F_1$-scores of the three baselines at a fixed number of tokens annotated for three datasets. When compared over the same number of tokens annotated, RS is a clear winner compared to PAS and LSS. Both LSS and PAS are likely to choose longer sentences; therefore, more tokens are annotated. Even though PAS contains some slight normalization which seems to help in some cases compared to LSS, it is not as good as RS when the cost is measured with the annotated number of tokens.

### 6.2 Effect of focusing on positive tokens

When we compare the other active learning query strategies based on sentence-based cost evaluation, no single strategy outperforms all the others across all the datasets (Table 5). However, among all the methods, the proposed `total-pos` is either the best or the second-best across all the datasets and iterations (12 out of 12). Recall that `total-pos` uses the predicted positive set, while `total` uses all tokens in evaluating a sentence's uncertainty (See Section 4.1.2). These two metrics do not corporate token count and annotation cost information into their formulation with normalization. Therefore, they favor longer sentences and attain better performance in sentence-based evaluation. While in the sentence-based cost evaluation `total-pos` yields similar performances, when token count and the corresponding annotation cost are taken into account, clearly the `total-pos` outperforms `total` in the CoNLL-03 and BC5CDR datasets as seen in Figure 4. In the third dataset (NCBI-Disease), both approaches yield similar learning curves. The NCBI-Disease dataset contains only one type of positive token and among the three datasets, it is the one with the lowest percentage of positive tokens. This class imbalance may have led to difficulty in estimating the positive token clusters accurately. Overall, for the two datasets, the proposed `total-pos` seems to be a better alternative compared to `total`, providing evidence that focusing on positive tokens' uncertainties is a better strategy than focusing on all tokens. For instance, in the CoNLL-03 dataset, `total-pos` achieves the same $F_1$-score of 0.865 but with ∼2400 fewer tokens compared to `total`. Similarly, `total-pos` reduces the required number of annotated tokens by more than ∼800 to obtain a $F_1$-score of 0.8 in the BC5CDR corpus.

**Table 5.** The $F_1$-scores of the baseline methods (first three rows) and average $F_1$-scores of different sentence score aggregation strategies in the last four iterations before convergence. The percentages of sentences queried are provided for the corresponding iterations under the iteration number. The reported deviations are the corresponding *standard error of the mean*.

| AL | CoNLL-03 | | | | BC5CDR | | | | NCBI-Dz. | | | |
|---|---|---|---|---|---|---|---|---|---|---|---|---|
| | Iteration 8 | Iteration 9 | Iteration 10 | Iteration 11 | Iteration 6 | Iteration 7 | Iteration 8 | Iteration 9 | Iteration 6 | Iteration 7 | Iteration 8 | Iteration 9 |
| Methods | %4 | %9 | %17.5 | %35 | %1.5 | %3 | %6 | %11 | %5 | %10 | %19 | %39 |
| RS | 76.5±0.2 | 79.0±0.8 | 81.7±2.5 | 84.7±3.4 | 66.9±0.5 | 69.9±0.5 | 73.7±0.3 | 79.1±0.1 | 61.1±1.1 | 64.0±0.9 | 69.5±0.5 | 78.2±0.2 |
| LSS | 66.3±0.4 | 70.1±0.5 | 73.6±0.4 | 74.3±0.6 | 62.6±0.4 | 65.2±0.7 | 73.4±0.3 | 79.8±0.1 | 51.8±0.7 | 61.6±0.6 | 71.3±0.5 | 78.6±0.3 |
| PAS | 65.9±0.9 | 69.5±0.8 | 82.7±0.3 | 87.2±0.2 | 54.6±1.8 | 59.4±1.0 | 66.2±0.8 | 75.1±0.8 | 57.5±0.7 | 61.7±0.8 | 69.7±0.9 | 78.2±0.4 |
| $\mu(\texttt{single})^a$ | 77.5±0.4 | 84.9±0.3 | 90.1±0.1 | 92.6±0.1 | 70.8±0.4 | 72.8±0.4 | 78.2±0.2 | 82.6±0.1 | 60.6±1.4 | 68.4±1.1 | 76.6±0.6 | 81.5±0.3 |
| $\mu(\texttt{total})^b$ | 78.3±0.4 | 86.5±0.2 | 90.8±0.1 | 92.8±0.1 | 69.0±0.6 | 75.2±0.4 | 80.3±0.1 | 83.5±0.1 | 60.2±1.1 | 70.9±0.7 | 76.7±0.6 | 81.3±0.4 |
| $\mu(\texttt{normalized})^c$ | 63.8±1.2 | 75.0±0.7 | 85.4±2.6 | 90.1±2.6 | 69.2±0.6 | 71.6±0.4 | 76.8±0.3 | 82.0±0.1 | 54.8±2.1 | 60.0±1.3 | 72.5±0.6 | 79.9±0.4 |
| $\mu(\texttt{total-pos})^d$ | 79.2±0.4 | 86.4±0.2 | 90.5±0.2 | 92.6±0.1 | 69.7±0.5 | 74.4±0.4 | 79.5±0.2 | 83.0±0.2 | 60.7±1.2 | 69.8±1.0 | 76.8±0.6 | 81.3±0.3 |
| $\mu(\texttt{dnorm-pos})^e$ | 75.7±0.5 | 84.3±0.2 | 89.8±0.1 | 92.4±0.1 | 70.8±0.5 | 73.0±0.5 | 77.2±0.3 | 82.0±0.1 | 57.2±1.9 | 66.2±1.0 | 75.3±0.6 | 81.1±0.4 |

[a] Average $F_1$-score of sTE, sTP, sTM, and sAP.
[b] Average $F_1$-score of tTE, tTP, tTM, and tAP.
[c] Average $F_1$-score of nTE, nTP, nTM, and nAP.
[d] Average $F_1$-score of tpTE, tpTP, tpTM, and tpAP.
[e] Average $F_1$-score of dpTE, dpTP, dpTM, and dpAP.

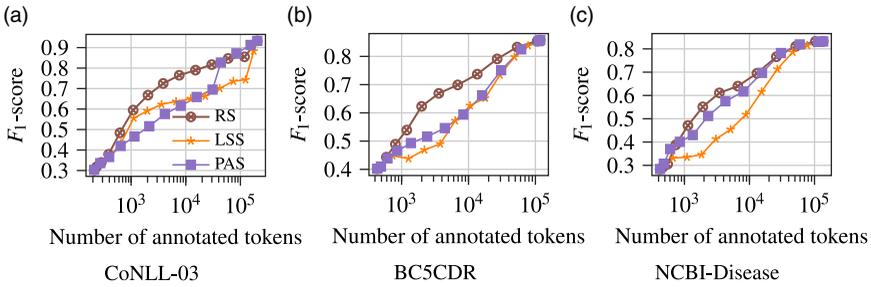

**Figure 3.** Average $F_1$-scores of RS, LSS, and PAS methods with respect to the total number of annotated tokens.

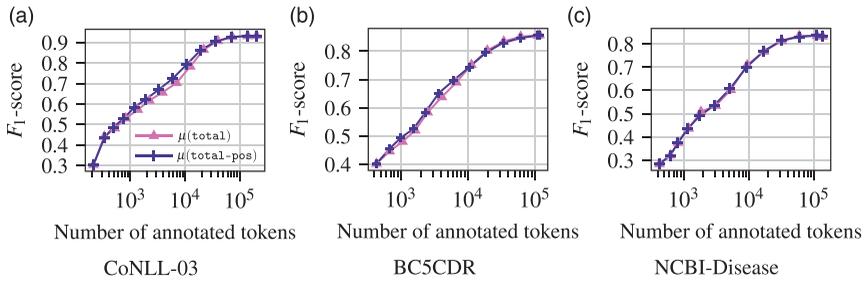

**Figure 4.** Average $F_1$-scores of `total` and `total-pos` methods with respect to the total number of annotated tokens.

### 6.3 Comparison of different sentence length normalization techniques

Of the different sentence normalization cases, `dnorm-pos` consistently performs better than the `normalized` approach when the number of sentences is the cost metric (Table 5). Thus, normalizing using positive tokens and using the sentence length density distribution is a much better approach than merely normalizing based on the number of tokens. We should also note that `normalized` consistently yields the lowest scores among all five active learning strategies, indicating that sentence normalization is not a good strategy at all. Figure 5 also displays the performance of these two strategies with normalization with respect to the number of annotated tokens. According to the Figure 5, `dnorm-pos` consistently outperforms `normalized` in CoNLL-03 dataset and slightly this time in NCBI-Disease dataset. Although `normalized` is marginally ahead for a few iterations, the two are comparably in the BC5CDR dataset. Overall, these results indicate that `dnorm-pos` is better than `normalized` for both cost measures. Similarly, `dnorm-pos` obtains the same $F_1$-score of 0.84 with `normalized`, but reduces the number of annotated tokens by ~3300 in the CoNLL-03 dataset. Similarly, `dnorm-pos` achieves the same $F_1$-score of 0.75 with ~1500 fewer annotated tokens compared to `normalized` in the NCBI-Disease corpus.

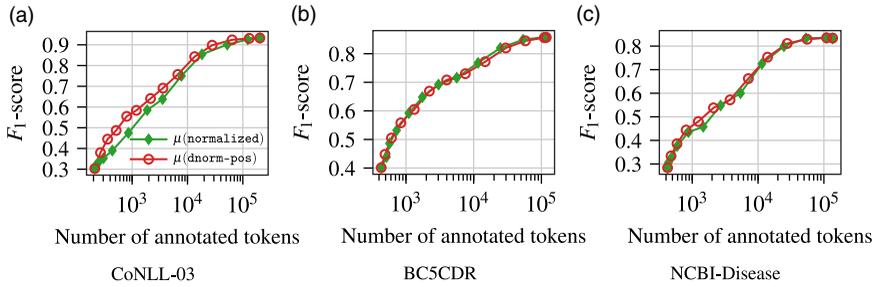

**Figure 5.** Average $F_1$-scores of `norm` and `dnorm-pos` methods with respect to the total number of annotated tokens.

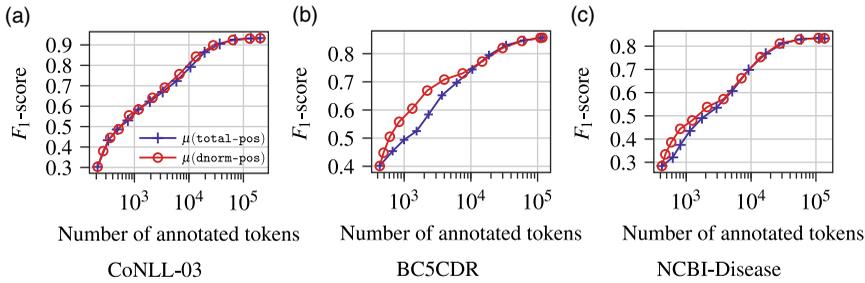

**Figure 6.** Average $F_1$-scores of `total-pos` and `dnorm-pos` methods with respect to the total number of annotated tokens.

Finally, we compare the two proposed approaches: `dnorm-pos` and `total-pos`. When the sentence is the evaluation metric, as seen in Table 5 `total-pos` either performs similarly or slightly better than `dnorm-pos`, possibly because it selects longer sentences. However, when we compare them in terms of token-annotated cost, `dnorm-pos` reaches the same level of $F_1$-scores with few token annotations as seen in Figure 6.

### 6.4 Overall comparison

Table 6 lists the approaches that attain a specified $F_1$-score with the least number of annotated tokens or sentences for a given uncertainty measure. Since the $F_1$-score depends on the dataset, different $F_1$-score values are picked for each dataset. The methods that use the least number of annotated sentences are shown in bold, and the method that reaches the predefined $F_1$-score level first with the least number of annotated tokens is also shown in italics. According to Table 6, the aggregation method (along within the columns of the table) is usually consistent, but the best methods differ for datasets and for different $F_1$-score levels (along the rows of the table). This clearly indicates that aggregation methods have a huge impact on the performances of the AL query methods.

Another important observation is the consistently good performance of `dnorm-pos` across different uncertainty measures and datasets, see entries that start with "dp." The most notable performance improvement, achieved by `dnorm-pos`, is observed in the CoNLL-03 dataset, which has the highest positive annotation to negative annotation ratio among all datasets (Table 1). At higher $F_1$-score levels, `dnorm-pos` methods outperform all other AL methods and baselines both in the sentence and token cost metrics. In BC5CDR, `dnorm-pos` performs well; in many cases, it is not only cost-efficient in terms of annotated tokens, but it is also cost-efficient when the cost metric is the number of annotated sentences. Results on the NCBI-Disease dataset are rather mixed, but as seen in Figure S1, at higher $F_1$-scores, density-based normalization and positive-based scores dominate. Figure S1 in the Appendix presents the performance of all AL approaches and the Random Sampling approach with respect to the number of annotated tokens.

**Table 6.** Comparison of aggregation methods for each uncertainty measure.[a] The method which achieves the indicated $F_1$-score with the least number of tokens is reported. Proposed methods are italicized. In each column, the method that requires the least number of sentences to achieve the indicated $F_1$-score is stated in bold. The abbreviations are listed in Table 2 and Table 3.

| | CoNLL-03 | | | | BC5CDR | | | | NCBI-Dz. | | | |
|---|---|---|---|---|---|---|---|---|---|---|---|---|
| Uncertainty measure | 60.0 | 70.0 | 80.0 | 90.0 | 50.0 | 60.0 | 70.0 | 80.0 | 50.0 | 60.0 | 70.0 | 80.0 |
| Token prob. | sTP | sTP | **dpTP** | *dpTP* | *dpTP* | nTP | ***dpTP*** | nTP | sTP | sTP | **tTP** | nTP |
| Token margin | **sTM** | **sTM** | *dpTM* | *dpTM* | *dpTM* | nTM | nTM | nTM | **sTM** | *tpTM* | tTM | **sTM** |
| Token entropy | sTE | sTE | *dpTE* | ***dpTE*** | *dpTE* | ***dpTE*** | *dpTE* | nTE | nTE | **sTE** | sTE | *dpTE* |
| Assign. prob. | sAP | sAP | *dpAP* | *dpAP* | ***dpAP*** | nAP | nAP | **tAP** | *dpAP* | sAP | tAP | *dpAP* |

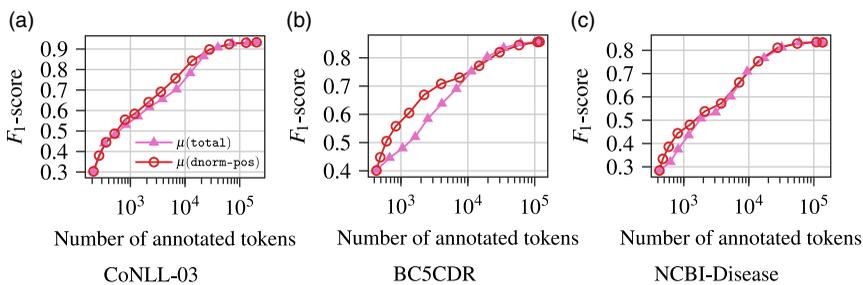

**Figure 7.** Average $F_1$-scores of `total` and `dnorm-pos` methods with respect to the total number of annotated tokens.

As discussed in Section 6.2, Table 5 shows that `total` methods tend to be the best-performing method when the evaluation is done with respect to the number of labeled sentences. On the other hand, `total-pos` methods and `dnorm-pos` methods perform comparably, but slightly worse than the `total` aggregation method. However, as seen in Figure 4, `total-pos` achieves the same $F_1$-score with a reduced number of annotated tokens in both CoNLL-03 and BC5CDR datasets. The improvement is even more significant in `dnorm-pos`. `dnorm-pos` can achieve the same $F_1$-score with `total` with a much less number of tokens, especially in the intermediate & late rounds of active learning on CoNLL-03, in the early and intermediate rounds on BC5CDR, and in the initial rounds on NCBI-Disease (Figure 7).

### 6.5 Computational cost

The cost of computing BERT embeddings can be neglected since they are computed once at the beginning and used as CRF feature values for each AL method. Similarly, the computational cost of the training is the same for all the AL methods.

Our approach involves semi-supervised dimensionality reduction, clustering, and outlier detection steps. UMAP might be considered as an expensive algorithm. Nevertheless, the time it takes to run the dimensionality reduction step is on the order of minutes for a single active learning iteration. Compared to the relatively expensive semi-supervised UMAP algorithm, the costs of running HDBSCAN and GLOSH are negligible, that is they take less than a minute on average for a single iteration in the datasets we tested. Overall, regardless of the extra cost, our approach stays feasible in terms of the runtime performance (see Figure S2).

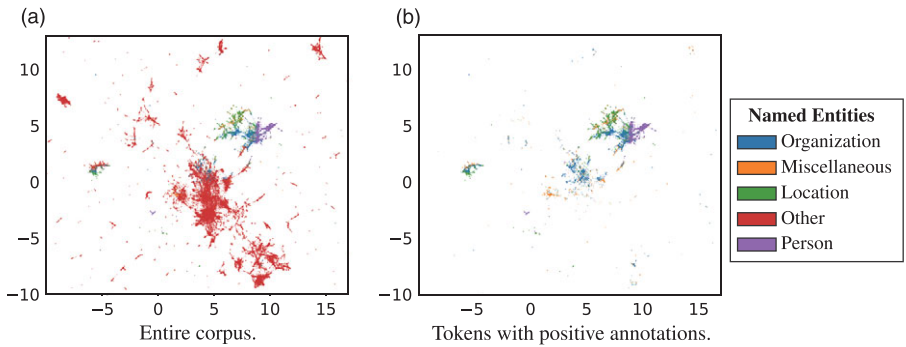

**Figure 8.** BERT embeddings of CoNLL-03 reduced to two dimension by semi-supervised UMAP, %2 labeled.

### 6.6 Performance of identifying positive token set in the query step

The query function relies on identifying the likely positive set before the classification step. We solve this step through clustering of tokens based on their BERT embeddings (as detailed in Section 4.1.1). As it can be seen from Table 1, the total number of sentences is often in the scale of thousands. However, even in the case of exponentially increasing batch size, the total number of queried sentences reaches 256 after the 8th query. Thus, clusters consist of unlabeled tokens until the very late active learning. Therefore, predicting the negative token set based on the majority classes (positive vs. negative) of each cluster is not reliable and does not lead to consistent assignments. Instead, we used a simpler and more robust approach by utilizing the imbalanced distribution of the number of classes and predicted the largest cluster as the negative token set. In this section, we inspect how well this step works. Figure 8 demonstrates an example of CoNLL-03 BERT embeddings reduced to two-dimensional space by applying semi-supervised UMAP with only %2 of labeled sentences (randomly selected). It can be seen from Figure 8a, that negative tokens form a large and dense cluster in the two-dimensional embedding space. We then calculate the precision and recall of the likely positive set. This step achieves 0.95/0.86/0.95 precision for negative tokens and 0.92/0.73/0.84 recall for the positive tokens for CoNLL-03/BC5CDR/NCBI-Disease datasets respectively at the first iteration without any labeled set. In contrast to the relatively high precision scores of the negative tokens, the precision scores of the positive tokens are 0.21/0.11/0.9, respectively. However, this is not a major problem for our end goal. Because the false-positive `other` annotated tokens are usually the stray data points and small outlier clusters, misclassification of them does not necessarily lead to missing/ignoring informative tokens. Furthermore, one can also argue that querying negative tokens whose embeddings differ from the rest of the negative tokens would be informative for the NER model.

We tune the hyper-parameters of the UMAP and HDBSCAN in the first iteration when there is no labeled set for computational efficiency. The performance of identifying a likely positive set could increase if they are tuned at every step. As an example, for CoNLL-03, we indeed achieved 0.98 precision for negative annotations and 0.97 recall for the positive annotations by using the method described above when only 1% of tokens are annotated, and hyper-parameters are re-tuned. However, since tuning is expensive, and the recall of positive tokens is quite high already; we did not do this. With larger computing resources this step could be further improved with hyper-tuning at each iteration.

As mentioned in the Sections 4.3 and 4.1.1, we use semi-supervised extension of UMAP. The quality of the low-dimensional embeddings tends to improve as the number of the labeled data instances increases at each step (Sainburg *et al.* 2021). For example, when %20 of sentences are annotated, we succeed in improving the precision scores of negative tokens from 0.86/0.95 to 0.91/0.96, respectively, for BC5CDR and NCBI-Disease. Similarly, the recall scores of the positive tokens increased from 0.73/0.84 to 0.74/0.88. Consequentially, initial $F_1$-scores 0.43/0.34/0.39 rose to 0.54/0.46/0.42, respectively for CoNLL-03, BC5CDR, and NCBI-Disease. However, we could not observe any further performance improvement with more labeled tokens, as opposed to our expectations. Therefore, in a setting with limited computational resources, one may prefer to use unsupervised UMAP and could compute low-dimensional embeddings only once instead of in every iteration.

Lastly, to assess the effect of the outlier detection method GLOSH on the performance, we conducted two different experiments for comparison. In the first one, we replaced the outlier detection GLOSH method with another method; we used the "Local Outlier Factor" (LOF) (Breunig *et al.* 2000). In the second experiment, we omit the outlier detection step. We report a representative result on the CoNLL-03 dataset, using the tpTE method. As shown in Table S2, results indicate that detecting outliers with GLOSH and eliminating them from the negative token set yields better performance compared to LOF, and the case where no outlier detection is conducted.

## 7. Conclusion

In this work, we focus on active learning for NER. One challenge of applying NER in active learning is the abundance of negative tokens. Uncertainty-based sentence query functions aggregate the scores of the tokens in a sentence, and since the negative tokens' uncertainty scores dominate the overall score, they shadow the informative, positive tokens. In this work, we propose strategies to overcome this by focusing on the possible positive tokens. To identify positive tokens, we use a semi-supervised clustering strategy of tokens' BERT embeddings. We experiment with several strategies where the sentence uncertainty score focuses on positive tokens and show empirically on multiple datasets that this is a useful approach. A second challenge of querying sentences with NER is related to the varying length of the sentences. Longer sentences that contain more tokens can bring more information at once; however, the annotation cost is higher. Normalizing sentences with the tokens they contain, on the other hand, yields in querying too short sentences. In this study, we proposed to normalize the scores such that sentences with the typical length for the dataset are queried more often. We evaluated the suggested methods both based on sentence and token-based cost analysis to validate our strategy. Overall, we believe the work presented here can support active learning efforts in the NER task.

**Supplementary material.** The supplementary material for this article can be found at https://doi.org/10.1017/S1351324923000165

**Acknowledgements.** This work was supported by a grant from Sabancı University under project number EPD-2020-13.

**Competing interests.** The authors declare none.